# Why We Don't Have AGI Yet

Peter Voss, AIGO.ai, Austin TX, USA, peter@aigo.ai
Mlađan Jovanović, Singidunum University, Belgrade Serbia, mjovanovic@singidunum.ac.rs

**Abstract** – The original vision of AI was re-articulated in 2002 via the term 'Artificial *General* Intelligence' or AGI. This vision is to build 'Thinking Machines' – computer systems that can learn, reason, and solve problems similar to the way humans do. This is in stark contrast to the 'Narrow AI' approach practiced by almost everyone in the field over the many decades. While several large-scale efforts have nominally been working on AGI (most notably DeepMind), the field of pure **focused** AGI development has not been well funded or promoted. This is surprising given the fantastic value that true AGI can bestow on humanity. In addition to the dearth of effort in this field, there are also several theoretical and methodical missteps that are hampering progress. We highlight why purely statistical approaches are unlikely to lead to AGI, and identify several crucial cognitive abilities required to achieve human-like adaptability and autonomous learning. We conclude with a survey of socio-technical factors that have undoubtedly slowed progress towards AGI.

## A Brief History of 'AGI'

Originally, the term 'Artificial Intelligence', coined by John McCarthy in 1955 [1], referred to machines or computer programs that can think, learn, and reason the way humans do – that have the general *cognitive* ability that we possess. At the time he expected this to be achieved within a few months!

As it turned out this was a ***lot*** harder, so over the years and decades the field of 'AI' morphed into 'Narrow AI' – solving just one (type of) problem at a time. An often overlooked but crucial detrimental consequence of this shift was that focus was now on 'achieving a particular goal via the ingenuity of the programmer or data scientist' rather than 'how do we build a system that ***has*** general intelligence'.

Therefore, AI's focus had shifted from ***having*** internal intelligence to utilizing external intelligence (the programmer's intelligence) to solve particular problems. This preoccupation with achieving specific narrow goals also had the undesirable side-effect of ignoring the importance of ***adaptation*** and ***agency***. Intelligent entities must be able to autonomously adapt to changing circumstances and goals. The entities are agents that **proactively** initiate actions in their surroundings [2].

By 2001 a number of AI researchers had independently concluded that the time was ripe to return to the original vision of 'AI' and decided to join forces to write a book on this subject [3]. In 2002, three of the authors (Ben Goertzel, Shane Legg, and Peter Voss) coined the term 'Artificial General Intelligence' for the book title.

AGI refers to creating (semi-)autonomous, adaptive computer systems with the general cognitive capabilities typical for humans. The ability to support abstraction, analogy, planning and problem-solving.

# External vs Internal Intelligence

| External/ Fixed | Internal/ Adaptive |
|---|---|
| Focus on **having / coding** knowledge and skills | Focus on **acquiring / learning** knowledge and skills |
| Domain-specific, relatively **fixed**, rule-based or statistical abilities. | Ongoing **cumulative**, general, **contextual**, **adaptive** learning. |
| **Externally** initiated improvements. | **Autonomy** and experience-based improvements. |

**Figure 1.** Comparison of external and internal intelligence.

## The Value of AGI

Imagine a computer system that has the general *cognitive* prowess and knowledge of a medical cancer expert. Now imagine a million copies of this chipping away at this scourge. Also consider the following advantages of artificial researchers and workers:

- ✓ Massively lower cost than humans performing cognitive tasks.
- ✓ Much better at communicating and sharing knowledge with each other – no egos to get in the way!
- ✓ Toiling away 365/24/7, much faster and with better concentration.
- ✓ Not having the various 'distractions' that we have – e.g. family, vacation, hobbies, etc.
- ✓ 'Photographic memory', and instant access to all online information.
- ✓ A much-improved ability to think logically, and to perform complex planning and reasoning.

Now apply these powers to the many challenges facing humanity: energy, environment, poverty, physical and mental wellbeing, aging, conflict resolution, governance, etc.

AGI promises to be able to significantly increase human flourishing.

## GPT (Generative Pretrained Transformer) as an attempt to achieve AGI

The recent success of transformer-based systems and other Large Language Models (LLMs) has given rise to speculations that these powerful systems may, with relatively minor enhancements, grow into AGI. This is extremely unlikely given the hard requirements for human-level AGI such as reliability [4], predictability [5], and non-toxicity [6]; real-time, life-long learning; and high-level reasoning and metacognition.

GPT-based systems are by definition 'generative' (they produce artificial content based on probabilities of co-occurrences; they make up stuff) and pre-trained (almost all of their knowledge is non-adaptive, not acquired interactively or in real-time). Due to their correlational nature LLMs fundamentally lack robust reasoning. Moreover, neural networks can perform poorly on previously learned tasks when exposed to new data and learning new tasks (known as catastrophic forgetting) [7].

Despite these limitations, properly curated and cross-validated LLM queries could be extremely valuable for helping with the difficult task of training AGIs; helping to provide the massive amount of general knowledge that they will need.

## Human-Like Cognitive Abilities – Cognitive AI

Cognitive AI describes machine systems able to **understand** language, use commonsense knowledge, **reason**, and **adapt** to unseen circumstances, similar to humans [8].

AGI assistants and researchers that can help us solve important (and not so important) problems will need to ***operate in the real world***, to have a deep understanding of life and science, to effectively communicate with us, to use our tools and systems, and to be able to learn and innovate. This requires a specific approach to building AGI, one that focuses on real-time, life-long ***conceptual*** learning and reasoning.

While such AGIs will be super-human in many ways, they will be constrained by having to operate with incomplete and often contradictory information, and limited time and resources to perform their tasks. On the other hand, they will not require human-level sense-acuity or dexterity. One could call this the Helen-Hawking model of AGI (Helen Keller/ Stephen Hawking) – AGI with human-level cognition but not overall human-level physical ability. They do, however, need to have some means of capturing and interacting with our 4D world. This could be accomplished via, for example, PC screen, keyboard, and mouse access. Beyond that, AGIs will also be excellent tool users, just like us.

We see that Cognitive AI is the clearest, and most definitive and direct path to AGI.

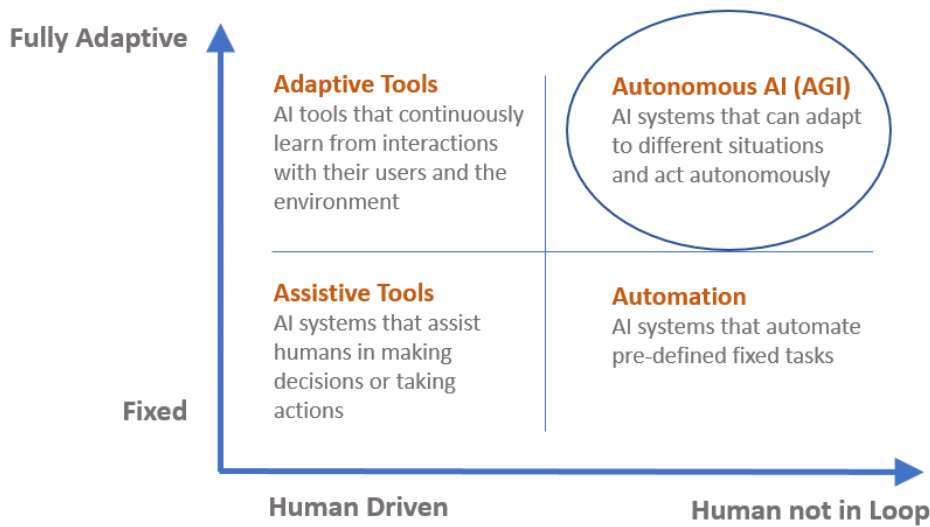

**Figure 2.** Dimensions of Adaptive Autonomous Intelligence.

## Defining the 'I' in AGI

Defining 'intelligence' as such is a thankless task, so instead, we will focus on a practical description of the kind of cognition/ intelligence required to achieve the AGI goals mentioned above. Essential requirements include:

- The ability to autonomously learn multi-dimensional objects, actions and sequences, including their appropriate context.
- To do this incrementally in real time, in an integrated manner (knowledge representation).
- To contextually identify known objects, actions, and sequences given partial or similar  input.

- The ability to generalize and to form new concepts and analogies autonomously.
- To learn to associate words with their respective concepts, and to learn language by understanding it.
- To learn action sequences and their associated 'triggers' via aping, exploration, trial-and-error, reinforcement, instructions, reasoning, and by other means (autonomous and semi-autonomous).
- The ability to adapt or unlearn existing knowledge and skills.
- The ability to reason abstractly, including planning, and theory-of-mind reasoning.
- Meta-cognitive reasoning and control (System 2 from [9]).
- Heuristic search and problem solving ability.
- To have and use both short-term and long-term memory for context, recognition, and reasoning.
- Means to focus on and select particular features available both externally and internally.

This list is not exhaustive but does cover the most essential cognitive abilities required for AGI. The ideas of AGI have been around for decades. But only recently, it has been recognized as an emerging technological megatrend [10].

## The Three Waves of AI

A few years ago DARPA presented a simple chronological taxonomy of AI called 'The Three Waves of AI' [11] broken up as follows:

1) Rule-based approaches also referred to as 'GOFAI' (Good Old-Fashioned AI), which dominated the field until about 2010 is the first 'wave'. This is characterized by largely hand-crafted data and algorithms. It includes expert systems, sophisticated logic and search algorithms, planning and scheduling systems, semantic web representation, natural language processing systems and the like. Its most visible successes were IBM's Deep Blue chess champion in 1997 and Watson, their Jeopardy quiz champion.

2) The second wave hit like a tsunami around 2012 when researchers figured out how to build neural networks using massive amounts of data and computational power, including GPU/TPUs. This led to breakthroughs in translation, image and speech recognition, mastery of many games (including Go) and ultimately powerful vision, speech, and text generation via GPTs. Currently, the pinnacle of these developments is represented by various LLMs such as ChatGPT. This wave is characterized by statistical and reinforcement learning; much of it is un- or self-supervised.

3) This final wave is still in its infancy. Its focus fully aligns with the requirements of AGI: Autonomous, real-time learning and adaptation, and high-level reasoning. It also expects concepts to be more grounded in reality (as opposed to language statistics), robust few-shot learning, and explainability. We would expect these systems to elegantly integrate sub-symbolic pattern matching with high-level symbolic and linguistic reasoning. An obvious candidate for all of these requirements is the 'Cognitive Architecture' approach.

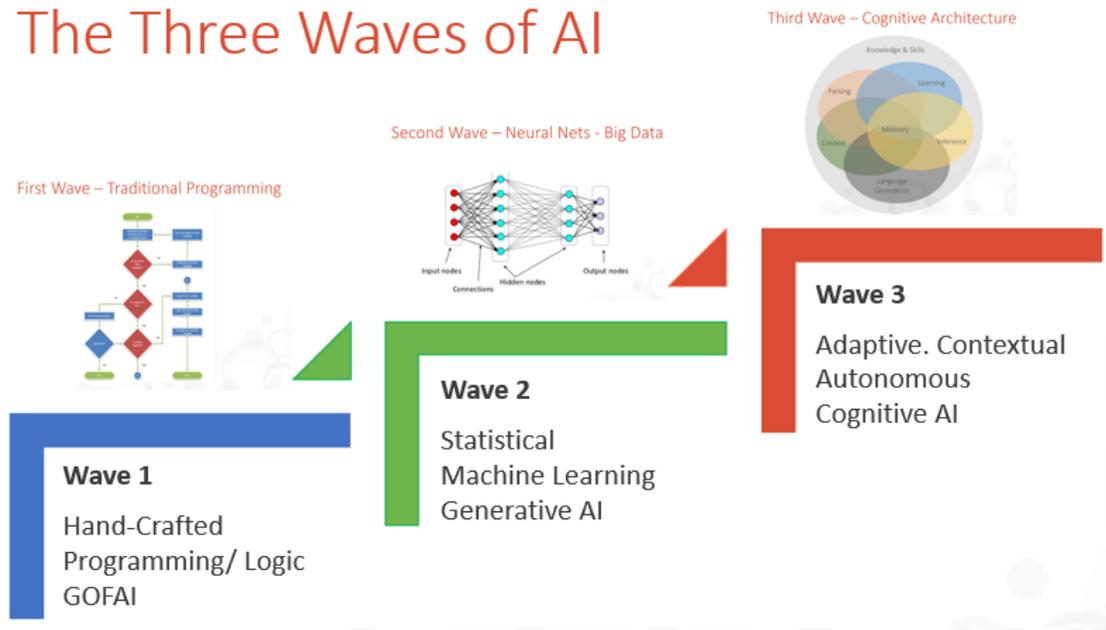

**Figure 3.** The Three waves of AI. Adapted from DARPA [11].

## Cognitive Architectures

Cognitive Architectures, or more generally 'Cognitive AI', are founded on the idea of creating systems that encompass and embody all of the essential structures required for a (human-level) mind. Importantly, it also considers how these structures and functions need to work together in conjunction with changing knowledge and skills to yield intelligence in diverse, dynamic environments [12].

Various cognitive architecture projects have been active for a few decades, though so far none have shown sufficient commercial promise to be widely adopted or particularly well-funded [8]. The reasons are manifold and complex (see next section), but a common theme is that they are implemented in far too modular and inefficient ways [13], and they lack a deep theory of learning and cognition [14].

A novel Cognitive Architecture designed to overcome these limitations is detailed in 'Concepts is All You Need: A More Direct Path to AGI' [15].

## Why don't we have AGI yet?

The short answer may well be that there simply hasn't been a project with the right approach/theory plus an adequate amount of funding. Recent success of ChatGPT suggests that hardware limitations may ***not*** be a major bottleneck at this time, making feasible highly sophisticated language production or 'inference'.

A longer answer includes the following considerations:

❖ Undoubtedly a major factor is that in spite of tens of thousands of AI researchers working in the field, only a tiny number are, by their own admission or by objective analysis, actually ***directly*** working on achieving AGI.

❖ One objective measure is whether the AI work done involves a clearly identified step or aspect of an overall detailed plan to achieve AGI. Very little AI work matches this criterion. Specifically, Generative AI research does not.

❖ Even projects dedicated to developing AGI are seldom implemented with an explicit theory that actually matches the requirements of the kind of autonomous, adaptive intelligence needed for AGI.

❖ Because of the tremendous success of Statistical AI (as opposed to Cognitive AI) over the past decade, currently almost all of the leading experts and practitioners in the field come from statistics, mathematics, or formal logic. This makes it almost impossible for them to see AGI requirements from a cognitive perspective.

❖ Unfortunately, motivations and incentives for individuals, teams, and companies are poorly aligned to optimizing progress towards AGI. Quite the contrary. For academics it is to publish rather than to develop. For companies it is to produce impressive demos, or to beat humans at some game or activity in order to secure additional funding. For most it is to beat existing benchmarks and not to change them.

❖ Using existing benchmarks for AGI is highly problematic: Firstly, focus on incremental improvements to specific existing benchmarks takes effort away from working on other problems that are actually more fundamental to achieving AGI. It is easier to work on things that you know how to make progress on, than to tackle difficult unknown issues. Secondly, current benchmarks are extremely poor at measuring progress of proto-AGIs. Early AGI systems will by definition do very poorly on existing narrow benchmarks, as well as on high-level IQ or professional admission tests.

❖ Even if all the stars are aligned in favor of developing AGI – a good theory and development plan, great cognitive team and funding, the right benchmarks – there still lurks what we can call 'The Narrow AI Trap'. Human nature is such that we instinctively want to show maximal progress in the shortest time. Unfortunately, for AGI this often means that we end up using external human intelligence to achieve a specific result or make progress on a given benchmark rather than implementing it in a way that puts the intelligence (adaptive, autonomous problem-solving ability) into the system. It takes careful discipline to avoid this. Naturally, AI efforts that are only nominally AGI (without good theory or plan), will more easily fall into this trap.

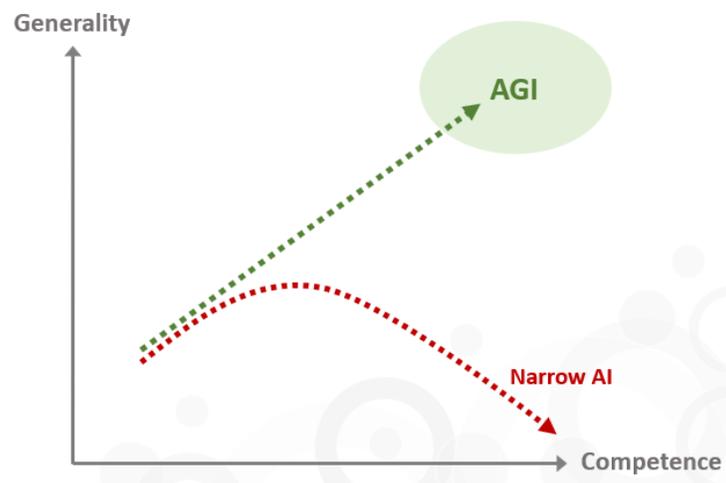

**Figure 4.** Pressure for near-term results and for beating existing non-AGI benchmarks tends to deflect projects away from AGI capabilities, towards narrower or commercial implementations.

- ❖ Finally, it is worth mentioning a lack of clear vision by people who could help to make AGI happen sooner. People who claim to not be motivated by money, but by helping mankind flourish. Yet many put their effort, reputation and money behind the latest fad or biggest potential short-term win.

## Conclusion

The spectacular performance of recent GPT technology teases the possibility that we may at last be close to being able to realize the original vision of 'AI' – to have human-level 'Thinking Machines'. The term 'AGI' was coined to (re)focus on this objective, and to bring about technology that can help us solve the many problems that humanity faces, to enhance human flourishing.

However, a detailed analysis of what human-level cognition requires shows that most of the technical approaches, motivations and benchmarks currently dominating the field of AI are ***not*** aligned with achieving this objective.

In order to accelerate progress towards AGI we will need to focus on the core requirements of human-like cognition – on items like autonomous, real-time, incremental learning; concept formation; and metacognitive control. We need to shift from Second to Third Wave AI, from Statistical or Generative AI to Cognitive AI.

## Acknowledgment

We thank Dr. Pat Langley for his constructive comments.